\pdfoutput=1

\documentclass[11pt]{article}

\usepackage[final]{lrec2026}
\usepackage{multirow}
\usepackage{booktabs}
\usepackage[margin=1in]{geometry}

\usepackage{latexsym}
\usepackage{amsmath}
\usepackage[T1]{fontenc}

\usepackage[utf8]{inputenc}
\usepackage{booktabs}
\usepackage{multirow}
\usepackage{adjustbox}
\usepackage{microtype}
\usepackage{float}
\usepackage{inconsolata}
\usepackage{pifont}
\usepackage{lineno}
\usepackage{graphicx}

\title{Enhancing Multi-Label Emotion Analysis and Corresponding Intensities for Ethiopian Languages}
\name{
\parbox{\textwidth}{
\centering
Tadesse Destaw Belay$^{1}$, Dawit Ketema Gete$^{2}$, Abinew Ali Ayele$^{3}$, \\ Olga Kolesnikova$^{1}$,
Iqra Ameer$^{4}$, Grigori Sidorov$^{1}$ and Seid Muhie Yimam$^{5}$
}
}

\address{$^{1}$Instituto Politécnico Nacional, Mexico City, Mexico, $^{2}$Wollo University, Kombolcha, Ethiopia, \\ $^{3}$Bahir Dar University, Bahir Dar, Ethiopia, $^{4}$Pennsylvania State University, PA, USA, \\ $^{5}$University of Hamburg, Hamburg, Germany\\
         tadesseit@gmail.com, \{tbelay23, kolesnikova, sidorov\}@cic.ipn.mx, userdavek@gmail.com,\\ abinewaliayele@gmail.com,  iqa5148@psu.edu, seid.muhie.yimam@uni-hamburg.de}

\abstract{
Developing and integrating emotion-understanding models are essential for a wide range of human-computer interaction tasks, including customer feedback analysis, marketing research, and social media monitoring. Given that users often express multiple emotions simultaneously within a single instance, annotating emotion datasets in a multi-label format is critical for capturing this complexity. The \textbf{EthioEmo} dataset, a multilingual and multi-label emotion dataset for Ethiopian languages, lacks emotion intensity annotations, which are crucial for distinguishing varying degrees of emotion, as not all emotions are expressed with the same intensity. 
We extend the EthioEmo dataset to address this gap by adding emotion intensity annotations. Furthermore, we benchmark state-of-the-art encoder-only Pretrained Language Models (PLMs) and Large Language Models (LLMs) on this enriched dataset. Our results demonstrate that African-centric encoder-only models consistently outperform open-source LLMs, highlighting the importance of culturally and linguistically tailored small models in emotion understanding. Incorporating an emotion-intensity feature for multi-label emotion classification yields better performance.  The data is available at \url{https://huggingface.co/datasets/Tadesse/EthioEmo-intensities}.
}

\begin{document}
\maketitleabstract
\section{Introduction}

Human emotion understanding is one of the most challenging and subjective tasks in Natural Language Processing (NLP) \cite{ziems-etal-2024-large}. Unlike many other NLP tasks, it requires assigning an emotion label(s) to a text that most accurately reflects the mental state of the author(writer) or a reader. The ability to detect emotions in text has numerous applications, from identifying (dis)satisfaction in customer feedback to evaluating the emotional well-being of individuals and societies \cite{10.3389/fpsyg.2023.1190326,ZHANG2024104835,pereira2024systematic}. How people convey their views and emotions is inherently diverse, often shaped by sociodemographic factors such as cultural background, personal experiences, communication styles, and emotional states \cite{hoemann2025construction}.

\begin{figure}[!h]
    \centering
    \includegraphics[width=\columnwidth]{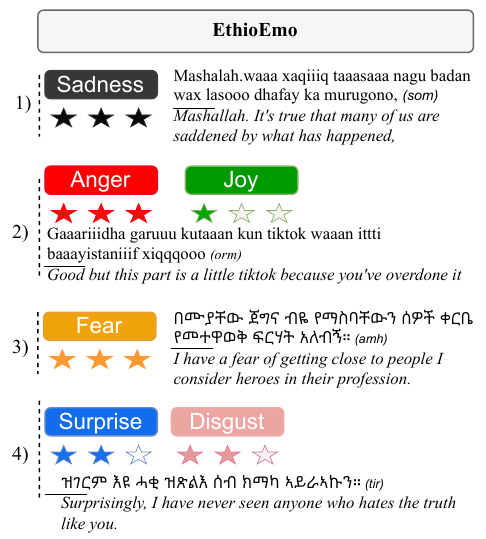}
    \caption{The EthioEmo dataset is a multi-label emotion dataset for Ethiopian languages, enhanced with new emotion intensity features. Each text may express one, two, several, or all emotions, and each annotated emotion is assigned an intensity level ranging from 1(low), 2(medium), or 3(high).}
    \label{fig:emoandint}
\end{figure}

It is critical to adopt systematic methodologies for organizing emotions in textual data in order to effectively analyze and interpret the complex and varied ways they are expressed. This involves employing structured annotation methods to categorize emotions and their intensity scales meaningfully \cite{zhang2024evaluation}.
There are two approaches in annotating an emotion dataset: \textbf{single-label} and \textbf{multi-label}. In the single-label approach, a text is assigned to either a single emotion class or no emotion. In contrast, the multi-label approach allows a text to be associated with none, one, multiple, or all of the targeted emotion labels. Both single-label and multi-label annotations can be multiclass, i.e., datasets with more than two classes.  Furthermore, emotion intensity is an extension of the emotion classification task that quantifies the strength or degree of each expressed emotion \citep{8282664}.  In emotion annotation, especially in multi-label emotion, adding the intensity of each corresponding emotion is very crucial, as each labeled emotion might not always be equally expressed in a content \cite{firdaus-etal-2020-meisd,labat-etal-2022-variation}.

Detecting the strength/intensity of emotion helps to understand its urgency and emphasis. For instance, some feelings may be subtly present, while others dominate more prominently. This complexity highlights the importance of assessing intensity, as it provides a nuanced understanding of how emotions are expressed. Consider a sentence example, \emph{`Although I'm incredibly excited about starting my new job, I feel a little sad about leaving my friends I made there.'} Here, the sense of happiness (joy) is pronounced and primary, whereas the feeling of sadness is secondary and less intense.  As illustrated in Figure \ref{fig:emoandint}, some texts have a single emotion label with its corresponding intensity scale value, while others have multiple emotions, each with its own intensity scale.

The work by \citet{belay-etal-2025-evaluating} created \textbf{EthioEmo}, a multi-label emotion dataset for four Ethiopian low-resource languages, namely Amharic (\texttt{amh}), Oromo (\texttt{orm}), Somali (\texttt{som}), and Tigrinya (\texttt{tir}). However, this multi-label emotion dataset is annotated without considering the intensity of each labeled emotion.
The main contributions of our works are: 
\begin{itemize}
    \item Extending the EthioEmo dataset to incorporate the multiple emotions along with their corresponding intensity to provide the complete affective information of a given instance, thereby enriching the applicability of the dataset for nuanced emotion and corresponding intensity analysis.
    \item Conducting a comprehensive evaluation of BERT-based encoder-only pre-trained language models (PLMs), open-source large language models (LLMs), and proprietary LLMs for their effectiveness in multi-label emotion classification and intensity prediction.
    \item Exploring the feasibility of cross-lingual transfer learning in a low-resource language setup and emotion transferability across the Ethiopian languages.
\end{itemize}

\section{Related Work}
\paragraph{Multi-label Emotion:} 
Emotion is central to human nature, and as online interactions grow, users express and react to a content in various ways.  A text expression is the major one and can simultaneously manifest multiple emotions to reflect the complex emotional nuances conveyed \citep{8282664}. To handle this complex multiple emotion expressions simultaneously, some of the recent multi-label emotion datasets are SemEval-2018 Task 1 \cite{mohammad-etal-2018-semeval}, GoEmotions \cite{demszky-etal-2020-goemotions}, EmoInHindi \cite{singh-etal-2022-emoinhindi}, WASSA-2024 Task 2 \cite{giorgi-etal-2024-findings}, BRIGHTER \cite{bright-muhammad2025}, EthioEmo \cite{belay-etal-2025-evaluating}, and SemEval2025 Task 11 data \cite{muhammad-etal-2025-semeval} are among them. While emotion recognition is widely studied, merely identifying the type of emotion in text is often insufficient for decision-making \cite{al2024challenges}; its corresponding intensity is very crucial.

\begin{table*}[!h]
    \centering
    \scalebox{0.95}{
    \begin{adjustbox}{width=2.2\columnwidth, center}
        \begin{tabular}{lllp{4.0cm} p{5.0cm}}
        \toprule
        {\bf Intensity dataset} &{\bf language(s)} &{\bf \# of isntance} & {\bf Emoiton classes} & {\bf Intensity lalebs} \\
        \midrule
        \citet{mohammad-etal-2018-semeval} & eng,ara,spn & 10,983/ 4,381/7,094& 12 emotin classes &1 (low), 2 (medium), 3 (high)  \\
        \citet{demszky-etal-2020-goemotions} & eng & 54.3k& 27 emotion categories &No intensities  \\
        \citet{ohman-etal-2020-xed}& eng,fin & 25k / 30k& 8 emotions + neutral&No intensities  \\
        \citet{firdaus-etal-2020-meisd}& eng &13k & 6 emotions + neutral &1 (low), 2 (medium), 3 (high)\\
        \citet{ciobotaru-etal-2022-red}&eng & 5,449 & 6 emotions + neutral&No intensities  \\
        \citet{singh-etal-2022-emoinhindi} & hin & 1,814&  15 emotion classes &1 (low), 2 (medium), 3 (high)  \\
        \citet{RAHMAN2024445} &eng & 6,037 & 8 depression emotions&No intensities  \\
        \citet{bright-muhammad2025} & 28 languages & 1,645 - 9,272& 6 emotion categories &1 (low), 2 (medium), 3 (high)  \\
        \citet{plisiecki2024predicting}& pol&10k&6 emotions &1 - 5 point scale \\
        \citet{belay-etal-2025-evaluating} & amh,orm,som,tir &  5,915/ 5,737/ 5,654/6,135& 6 emotions + neutral &1 (low), 2 (medium), 3 (high)  \\
        \bottomrule
        \end{tabular}
    \end{adjustbox}
    }
    \caption{Summarized multi-label emotion and intensities datasets related works.}
    \label{tab:related-work}
\end{table*}

\paragraph{Intensity in Multi-label Emotion:}
In addition to emotion classification, analyzing the degree of each emotion provides deeper insights, leading to more informed and effective decisions \citep{10409495}. Accurately annotating the intensity of each labeled emotion is essential for advancing the capabilities of language models, as it presents an additional challenge for nuanced emotion recognition. Most common multi-label emotion datasets \cite{mohammad-etal-2018-semeval, singh-etal-2022-emoinhindi, giorgi-etal-2024-findings, bright-muhammad2025} include intensity scales for the corresponding labeled emotion. However, the EthioEmo dataset is annotated in a multi-label annotation setting without specifying the intensity of each corresponding emotion.
Inspired by this work and the importance of incorporating intensity in multi-label emotion annotation, we extend the EthioEmo dataset by adding an intensity feature. 
Related work on multi-label emotion and emotion-intensity datasets is summarized in Table \ref{tab:related-work}.

\paragraph{Cross-Lingual Experimentation:}
Cross-lingual transfer learning has emerged as a promising approach to address data scarcity in low-resource languages \cite{maladry-etal-2024-findings}. It has been used to transfer knowledge from high-resource to low-resource and among low-resource languages \cite{zhang-etal-2024-enhancing}. By utilizing cross-lingual approaches, one language can benefit from the resources and insights of another, thus enhancing model generalization over emotion-related tasks \citep{zhu-etal-2024-model,kadiyala-2024-cross,cheng-etal-2024-teii}. Cross-language experimentation could explore whether emotion classification can be improved by transferring knowledge across languages. \citet{navas-alejo-etal-2020-cross} explored various cross-lingual strategies for emotion detection and intensity grading, illustrating how models can adapt across different languages. However, the evaluation of cross-lingual transfer across different languages spoken within the same country has not been extensively studied. In this work, we conduct cross-lingual emotion analysis among Ethiopian languages, which are characterized by distinct script systems: Amharic (\texttt{amh}) and Tigrinya (\texttt{tir}) use the Ethiopic (Ge'ez) script, while Oromo (\texttt{orm}) and Somali (\texttt{som}) use the Latin script. 

\section{EthioEmo Dataset} \label{ethioemo}
The EthioEmo emotion dataset is an emotion dataset that covers four Ethiopian languages. The data was collected from social media platforms such as  X(Twitter) and news portals and annotated in a multi-label setup. The targeted emotion classes are the six basic emotion classes (anger, disgust, fear, joy, sadness, and surprise). Text without an emotion label is assigned the neutral label "0". Each instance of the dataset is annotated by a minimum of three and a maximum of five annotators. The final emotion labels are determined through a majority vote (two or more votes for the three annotators and three or more votes for the five annotators).

\paragraph{Emotion Intensity Annotation}
The degree of feeling in an emotion dataset is crucial for accurately understanding complex emotions \cite{firdaus-etal-2020-meisd}. We enhanced the EthioEmo dataset by including annotations for the intensity of each identified emotion. Annotators were trained to assign an intensity label to each emotion category. We utilized the customized version of the POTATO annotation tool \cite{pei-etal-2022-potato} along with in-house annotation practices. Annotators are trained to assign an intensity label to each identified emotion category. We follow the emotion intensity scaling approaches from previous works \cite{mohammad-etal-2018-semeval,singh-etal-2022-emoinhindi,muhammad-etal-2025-semeval} and the intensity scale comprises four levels: 0 (no intensity for any emotion class — neutral), 1 (slight emotion, e.g., slight anger), 2 (moderate emotion, e.g., moderate anger), and 3 (high emotion, e.g., very anger). Following the original dataset’s annotation setup and based on annotator availability, each instance in \texttt{orm}, \texttt{som}, and \texttt{tir} was annotated by a minimum of three annotators, while each instance in \texttt{amh} is annotated by five annotators, and the final label was determined by majority vote. 
The rationale for assigning a minimum of three annotators is based on previous annotations of related datasets \cite{belay-etal-2025-evaluating,mohammad-etal-2018-semeval,muhammad-etal-2025-semeval}. It also represents the minimum cost-effective number required to obtain a majority vote, particularly in settings with limited annotator availability. Annotation guidelines and details of annotators are found in Appendix \ref{app:guideline}.

The final intensity score for each emotion is obtained using the following aggregation rule. For three annotators per instance, at least two annotators must assign a non-zero intensity value (1, 2, or 3); instances with fewer than two non-zero annotations are discarded, see rule \ref{eq:three_annotators}. In the five-annotator setting (rule \ref{eq:five_annotators}), the rule is applied whenever at least two annotators assign a non-zero intensity. The intensity of each emotion for the three- and five-annotator settings is then computed using the following formulas.

For three annotators per instance \eqref{eq:three_annotators}:
\begin{equation}
\label{eq:three_annotators}
I_{\text{final}} =
\begin{cases} 
0, & \text{if } 0 \leq \text{Avg} < 1  \\
1, & \text{if } 1 \leq \text{Avg} \leq 1.5 \\
2, & \text{if } 1.5 < \text{Avg} \leq 2.5 \\
3, & \text{if } \text{Avg} \geq 2.5
\end{cases}
\end{equation}

For five annotators per instance \eqref{eq:five_annotators}:
\begin{equation}
\label{eq:five_annotators}
I_{\text{final}} = 
\begin{cases}
0 & \text{if } \text{Avg} \leq 1.5 \text{ and } \text{anno} \geq 2 \\
1 & \text{if } \text{Avg} \in [0.6, 1.5) \text{ and } \text{anno} \geq 2 \\
2 & \text{if } \text{Avg} \in [1.5, 2.5) \text{ and } \text{anno} \geq 2 \\
3 & \text{if } \text{Avg} \geq 2.5 \text{ and } \text{anno} \geq 2 \\
\end{cases}
\end{equation}

where $\text{Avg}$ is the average intensity score among the annotators of an instance.


\paragraph{Annotators Agreement} We obtained moderate results of inter-annotator agreement (IAA) based on Cohen's Kappa (see Table \ref{tab:emotion-scores}), where scores $<0.0$ indicate poor agreement, $0.00$--$0.20$ slight, $0.21$--$0.40$ fair, $0.41$--$0.60$ moderate, $0.61$--$0.80$ substantial, and $0.81$--$1.00$ almost perfect \cite{SnchezVelzquez2016LetsAT}.
The moderate intensity IAA agreement score shows the difficulty of the emotion’s intensity annotation task. 

\begin{table}[!h]
\centering
\resizebox{\columnwidth}{!}{%
\begin{tabular}{lccccccc}
\hline
\textbf{Lang.} & \textbf{Ang.} & \textbf{Disg.} & \textbf{Fear} & \textbf{Joy} & \textbf{Sadn.} & \textbf{Surp.} & \textbf{Avg.} \\
\hline
\texttt{amh} & 0.60 & 0.59 & 0.58 & 0.59 & 0.54 & 0.52 & \textbf{0.57} \\
\texttt{orm} & 0.52 & 0.50 & 0.48 & 0.53 & 0.50 & 0.53 & \textbf{0.51} \\
\texttt{som} & 0.59 & 0.47 & 0.48 & 0.49 & 0.50 & 0.41 & \textbf{0.49} \\
\texttt{tir} & 0.55 & 0.54 & 0.52 & 0.51 & 0.53 & 0.53 & \textbf{0.53} \\
\hline
\end{tabular}
}
\caption{Emotion intensity Cohen's Kappa \cite{cohen1960coefficient} inter-annotator agreement (IAA) scores across languages and emotions.}
\label{tab:emotion-scores}
\end{table}

\begin{figure*}[th!]
    \centering
    \includegraphics[width=\linewidth]{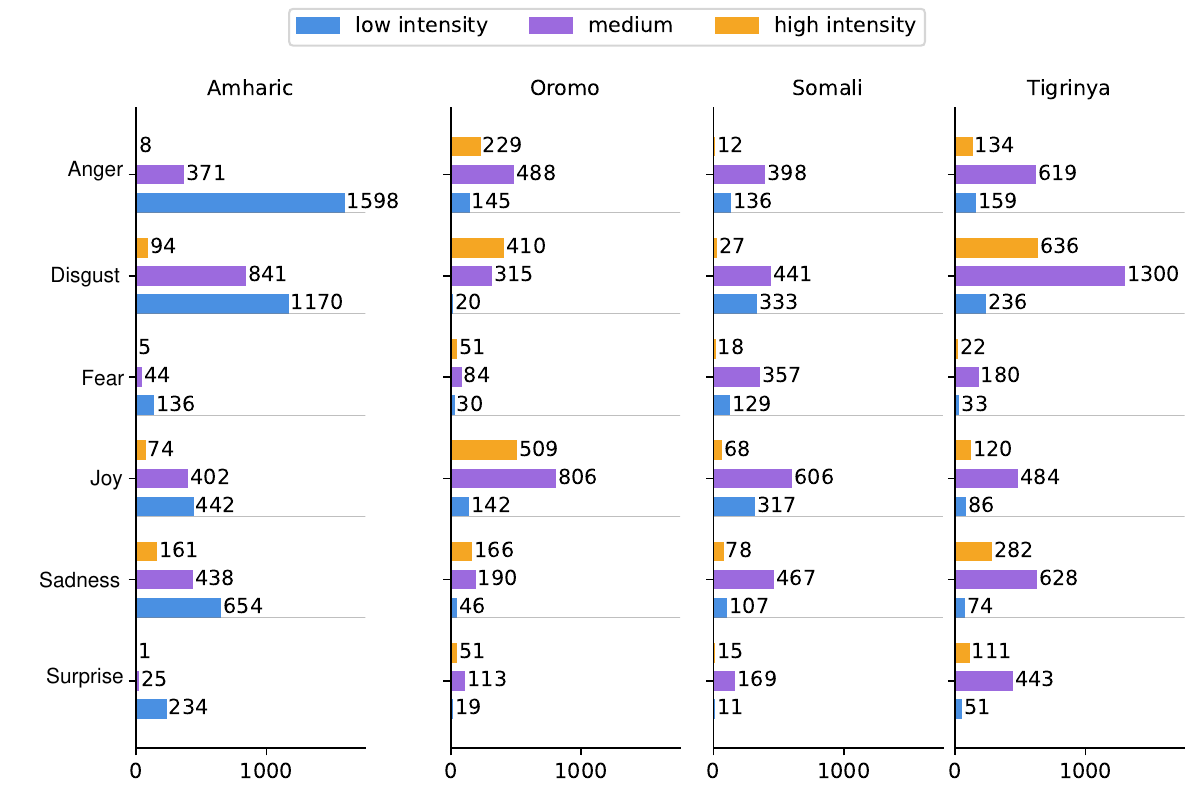}
    \caption{Emotion intensity distribution across emotion labels with three intensity levels (low, medium, and high of the corresponding emotion). Instances that have not been labeled in any of the given emotions are not included in the statistics, instances that have none of the targeted emotions are for Amharic (\texttt{amh}) is 1021, Oromo (\texttt{orm}) 1357, Somali (\texttt{som}) 2156, and Tigrinya (\texttt{tir}) 1336. The statistics of each emotion across all languages can be found in Appendix \ref{app:label-dist}.}
    \label{fig:intens-stat}
\end{figure*}

\paragraph{Emotion Intensity Distribution}\label{app:intes-dist} 
The distribution of the emotions and intensities in the dataset is presented in Figure~\ref{fig:intens-stat}.  Most instances have low or medium emotion intensity, while \texttt{Joy} in \texttt{orm}, \texttt{Disgust} in \texttt{tir}, and \texttt{orm} languages have many instances with high emotion intensity compared to other emotion classes. 

\section{Experiment Setup}
\subsection{Model Selection}
We select language models for evaluation from different perspectives, such as general multilingual PLMs,  Africa-centric PLMs, open-source LLMs, and proprietary LLMs. The rationale behind choosing LLMs is that small and large variants (such as Llama-3.1-8B and Llama-3.3-70B) focus on multilingual support and popularity.

\paragraph{General Multilingual PLMs} We evaluate the most common multilingual BERT-like PLMs such as LaBSE \cite{feng-etal-2022-language}, RemBERT \cite{chung2020rembert}, XLM-RoBERTa \cite{xlm-r}, mBERT \cite{libovický2019lmbert}, and mDeBERTa \cite{he2021debertav3}. 

\paragraph{Africa-centric PLMs}
We experiment with fine-tuning the most common African-centric language models such as AfriBERTa \cite{ogueji-etal-2021-small}, AfroLM \cite{dossou-etal-2022-afrolm}, AfroXLMR (61 and 76 languages) \cite{alabi-etal-2022-adapting}, EthioLLM \cite{tonja-etal-2024-ethiollm}, and AfroXLMR-Social \cite{belay2025afroxlmr}.

\paragraph{Open source LLMs}
Based on their popularity in the open-source community and better multilingual support, we evaluate the following open-source LLMs: Gemma-3-12B \cite{team2025gemma}, Llama-3.1-8B \cite{dubey2024llama}, Llama-3.3-70B \cite{dubey2024llama}, and DeepSeek-R1-70B \cite{deepseekai2025}. 

\paragraph{Proprietary LLMs} We examine the lightweight and latest versions of the proprietary models, GPT-4.1-mini \cite{openai} and Gemini-2.5-pro-flash \cite{gemini25} for reproducibility and cost-effectiveness.

\subsection{Evaluation Setup} 
For encoder-only models, we fine-tune them using the train–test split of the dataset for emotion classification, emotion intensity prediction, and cross-lingual transfer experiments. The training experiment settings are presented in Appendix \ref{app:parms} for reproducibility. 
For LLMs, Chain-of-Thought (CoT) prompting is a widely used strategy, particularly effective across various NLP tasks, including mathematical reasoning. However, prompting models to select from a predefined list of emotions, such as prompting LLMs to choose the emotion of a given text from [anger, disgust, fear, joy, sadness, surprise], often leads to over-prediction, where models assign multiple emotion classes even when most instances contain only a single emotion. To mitigate this, we employed CoT prompting \cite{cotprompt} with binary emotion prediction, evaluating one emotion at a time with a yes/no response.

\subsection{Formulating Evaluation Tasks} 
Using the aforementioned language models and the multi-label emotion dataset with corresponding intensity scales, we conduct experiments with a 60:10:30 train–dev–test split from a total datsets \texttt{amh} - 5,915, \texttt{orm} – 5,737, \texttt{som} –  5,654, and \textit{tir} –  6,135.
Using these datasets, we evaluate the following emotion analysis tasks:
\begin{itemize}
    \item \textbf{Multi-label emotion classification}: The emotion classification task is formulated as a binary decision for each emotion (one emotion at a time), where the model provides a yes/no (1/0) response to questions such as “\textit{Does the text convey {anger} or not?}” The same setup is applied for all other emotions.
    \item \textbf{Emotion intensity prediction}: For intensity prediction, we provide the text along with list of emotions and ask the model to predict the intensity level of each corresponding emotion from 0 (no), 1(low),  2 (medium), or 3 (high).
    \item \textbf{Cross-lingual multi-label emotion evaluation}: Our cross-lingual experiment setup involves two types: 1) fine-tuning BERT-like encoder-only models on datasets from all available languages except the target language, which is held out for evaluation, and 2) fine-tuning a single multilingual model that includes all the languages.
\end{itemize}


\subsection{Evaluation Metrics}
Based on the multi-label task evaluations from previous similar works \cite{mohammad-etal-2018-semeval,muhammad-etal-2025-semeval} and for a highly imbalanced data, we used the Macro-F1 score 
that takes the average of individual emotion F1 scores. For intensity prediction, we used the Pearson correlation coefficient (\textit{r}) \cite{schober2018correlation} between predicted and true intensity values.

\begin{table*}[!h]
    \centering
    \resizebox{\textwidth}{!}{%
    \begin{tabular}{lccccc||ccccc}
        \toprule
&\multicolumn{5}{c||}{\textbf{Multi-Label Emotion Classification (F1)}} & \multicolumn{5}{c}{\textbf{Intensity Prediction ( Pearson \textit{r})}}\\
\cmidrule{2-6}\cmidrule{7-11}
\textbf{Models}& amh& orm & som & tir & Avg. & amh& orm & som & tir & Avg.\\
\midrule
\multicolumn{6}{l}{\textit{General multilingual PLMs} }  \\
LaBSE  &66.51  & 41.49  & 43.99 & 48.88  & 50.22  
&47.79 &16.53 &25.70 & 32.10  & 30.53 \\
RemBERT  &60.15  & 47.54  & 48.31 & 50.37  & 51.59 
&52.73 & 24.15&24.85 & 37.63  & 34.84 \\
mBERT   &26.51  & 40.32  & 27.01 & 25.72  & 29.89  
&00.00 & 17.88& 5.51 & 3.13   & 6.63 \\
mDeBERTa &53.43  & 32.84  & 36.86 & 41.73  & 41.22  
&33.07 & 7.27 &7.02 & 19.24   & 16.15  \\
XLM-RoBERTa &63.73  & 37.42  & 33.51 & 13.32  & 37.00 
&53.63 &17.34&18.39 &15.95& 26.33\\
\midrule
\multicolumn{6}{l}{\textit{African-centric PLMs} }  \\
EthioLLM  &58.68  & 47.95  & 33.84 & 44.78  & 46.31  
&41.90 &21.58&9.96  &22.77& 24.05 \\
AfriBERTa &60.64  & 54.10  & 44.66 & 47.97  & 53.34   
&39.38 &25.24& 20.63&27.56& 28.20\\
AfroLM    &54.76  & 42.21  & 32.77 & 38.60  & 42.09  
&37.75 & 15.90&5.08 &18.42& 19.25 \\
AfroXLMR-61L &67.93 & 51.73  & 49.31 & 54.96  & 55.98 &55.19&26.75&37.81& 41.96 &40.43\\
AfroXLMR-76L&68.46 & 49.68  & 49.25 & 53.08  & 55.12
&\textbf{55.57}& 29.15 & 41.36 & 40.32  & 41.60  \\
AfroXLMR-Social & \textbf{70.66} & \textbf{60.74}  & \textbf{54.75} & \textbf{60.24}  & \textbf{61.60}
&53.82& \textbf{32.26} & \textbf{38.44} & \textbf{42.18}  & \textbf{41.68}  \\
\midrule
\multicolumn{6}{l}{\textit{Zero-shot - open-source LLMs} }  \\

 Llama-3.1-8B & 32.06  & 07.77 & 07.13 & 11.84  & 14.46 
 &14.58 &07.13 &09.36 &07.30 & 09.59 \\ 
 Gemma-3-12B & 42.19  & 23.28 & 32.05 & 32.57  & 32.62 
 &30.45&16.60 & 26.08&\textbf{22.17} &23.83 \\ 
DeepSeek-R1-70B& 36.89  & 28.15  & 26.56 & 26.49  & 29.52 
& 31.05&25.17 &26.26 & 21.78& 26.07 \\ 
Llama-3.3-70B & \textbf{42.84}& \textbf{29.84} &\textbf{32.49}&\textbf{32.93} & \textbf{34.53} &\textbf{39.52} &\textbf{27.31} &\textbf{30.08} &21.12 & \textbf{29.51}\\ 
\midrule
\multicolumn{6}{l}{\textit{Zero-shot - commercial LLMs} }  \\
Gemini-2.5-flash  & 24.56   & 24.11  & 16.38 & 12.31   & 19.34 
&24.51 &14.08 &11.16 &9.46 &14.80 \\ 
GPT-4.1-mini(0-shot)  & \textbf{46.73}   & 44.06  & 45.07 & \textbf{34.77}   & 42.66 
& 40.01&37.20 &41.49 &\textbf{29.71} &37.35 \\ 
GPT-4.1-mini(5-shot)  & 45.68   & \textbf{46.60}  & \textbf{48.10} & 32.73&\textbf{43.28}   & \textbf{41.21} & \textbf{37.59} &\textbf{45.94} &28.85 &\textbf{38.40} \\

\bottomrule
\end{tabular}
}
\caption{Multi-label emotion prediction macro F1 results (left) and emotion intensity prediction (right) results using Pearson correlation. The best performance scores are highlighted in \textbf{bold}. All evaluated open-source LLMs are instructed versions.} 
\label{tab:emo-intens}
\end{table*}


\section{Experiment Results}
\subsection{Multi-Label Emotion Classification}
The results of the multi-label emotion classification are presented in Table \ref{tab:emo-intens}. As shown, BERT-like encoder-only models achieve better performance than zero-shot LLMs.  AfroXLMR-Social achieves stronger results, possibly due to three main reasons: (1) it is a continual pre-tarined model from African-centric AfroXLMR model using domain specific social media corpus - a corpus from X (Twitter) and news, (2) it is based on multilingual XLM-RoBERTa that covers 100 languages \cite{xlm-r} and further fine-tuned on 76 African languages, and (3) it benefits from a larger parameter size (comparatively across encoder-only models) and more diverse training data than EthioLLM. AfroXLMR, the base of AfroXLMR-Social, was trained on approximately $\approx\!12$ GB of multilingual African text, which enables more effective cross-lingual transfer than EthioLLM (trained on $\approx\!3$ GB corpus). 

Large language models (LLMs) perform less effectively for the evaluated low-resource languages, and their performance is highly dependent on parameter size. For example, Llama-3.1-8B performs the worst among evaluated LLMs, while Llama-3.3-70B performs better. 
Comparably, Amharic is better represented than other Ethiopian languages. Overall, encoder-only models continue to outperform both open-source and proprietary LLMs in the multi-label emotion analysis for low-resource languages. African-centric PLMs are better at classifying multi-label emotions than LLMs.

\paragraph{How does adding corresponding intensity features enhance multi-label emotion performance?}
We evaluate our best encoder-only model, AfroXLMR-Social, to assess the impact of incorporating an emotion intensity feature for multi-label emotion classification. The model achieves macro F1 scores of 82.13 for \texttt{amh}, 62.36 for \texttt{orm}, 57.53 for \texttt{som}, and 65.19 for \texttt{tir} language. These results represent improvements ranging from 1.62 to 11.47 points compared to the baseline multi-label emotion-only results that are shown in Table \ref{tab:emo-intens}. Training with emotion intensity annotations provides a denser supervision signal than binary emotion-only labels. In particular, the model learns to emphasize high-intensity emotional signals while naturally reducing the ambiguity associated with low-intensity or borderline cases.

\begin{table*}[!h]
\centering
\resizebox{\textwidth}{!}{%
\begin{tabular}{lccccc||ccccc}
\toprule
&\multicolumn{5}{c||}{\textbf{Cross-lingual results (F1)}} & \multicolumn{5}{c}{\textbf{Cross-lingual with same script (F1)}}\\
\cmidrule{2-6}\cmidrule{7-11}
\textbf{Models}& amh& orm & som & tir & Avg. & amh& orm & som & tir & Avg.\\
\midrule
LaBSE  &44.11 & 20.77  & 35.18 &40.13  &35.55 &43.83&21.60&21.09&36.99&30.88\\
RemBERT &42.65 & 20.87  & 31.32 &33.39  &31.81 &37.81&33.90&23.11&12.31&26.78\\
mBERT &25.10 & 10.79  & 14.13 &18.27  &17.07 &28.15&23.51&18.10&19.95&22.43\\
mDeBERTa &36.40 & 26.63  & 18.83 &38.03  &29.97 &38.23&24.53&13.92&33.05&27.43\\
XLM-RoBERTa &23.52 & 23.69  & 26.98 &38.63  &28.21 &31.65&21.63&09.22&21.56&21.02\\
EthioLLM &38.37 & 22.46  & 22.76 &33.08  &30.42 &31.31&20.90&10.08&27.40&22.42\\
AfriBERTa  &46.28 & 35.86  & 30.81 &38.05  &37.75 &38.02&27.09&26.75&31.37&30.81\\
AfroLM  &32.12 & 10.38  & 9.00  & 25.48 &19.25 &28.67&23.05&11.53&21.75&21.25\\
AfroXLMR-61L&56.41 & 43.24& 42.21 & 52.70 &48.64 &53.13&27.84&12.68&44.57&34.56\\
AfroXLMR-76L &56.65  & 45.01  & 41.24 & 53.39  & 49.07  &45.95&28.47&14.06&49.65&34.53\\
AfroXLML-Social &\textbf{60.22}  & \textbf{52.30}  & \textbf{45.72} & \textbf{56.00}  & \textbf{53.56}  &\textbf{59.28}&\textbf{41.44}&\textbf{37.00}&\textbf{53.65}&\textbf{47.84}\\
\bottomrule
\end{tabular}
}
\caption{Cross-lingual emotion prediction results. Train with all languages except the held-out language results (left) and train with only similar scrips (right) column. The best performance are highlighted in \textbf{bold}.}
\label{tab:cross-lingual}
\end{table*}

\subsection{Emotion Intensity Prediction}
Table \ref{tab:cross-lingual}  presents the results of the intensity prediction. As all Ethiopian languages are not included during pretraining, mBERT performs worse; the slightly better performance on \texttt{orm} and \texttt{som} might be because these languages use the Latin script and share some vocabulary. Similarly, in the emotion classification task, AfroXLMR-Social achieves the highest performance in intensity prediction. LLMs at intensity prediction are worse than the emotion classification task, this might be due to the subjectivity and complexity of emotion intensity prediction pose a greater challenge even for high-resource languages, such as English \cite{bright-muhammad2025}. The overall results show that understanding emotions and predicting intensities from text is challenging.
\paragraph{How do LLMs help for emotion and its intensity annotation for low-resource languages?} State-of-the-art LLMs, such as GPT, demonstrate close to human-level performance in generating high-quality emotion and intensity annotations for English \cite{2024-effectiveness,bagdon-etal-2024-expert}. However, for low-resource languages, their performance drops significantly - below 50\% for all Ethiopian languages . A closer examination of the GPT prediction outputs reveals that the model often attempts to translate the input text into English before predicting emotions and their corresponding intensities  (particularly for Ethiopic script languages such as \texttt{amh} and \texttt{tir}), despite explicit prompt instructions to avoid explanations or translations. Consequently, the model frequently outputs “no emotion” and “no intensity” for these languages.

\begin{figure*}[!h]
    \centering
    \includegraphics[width=\linewidth]{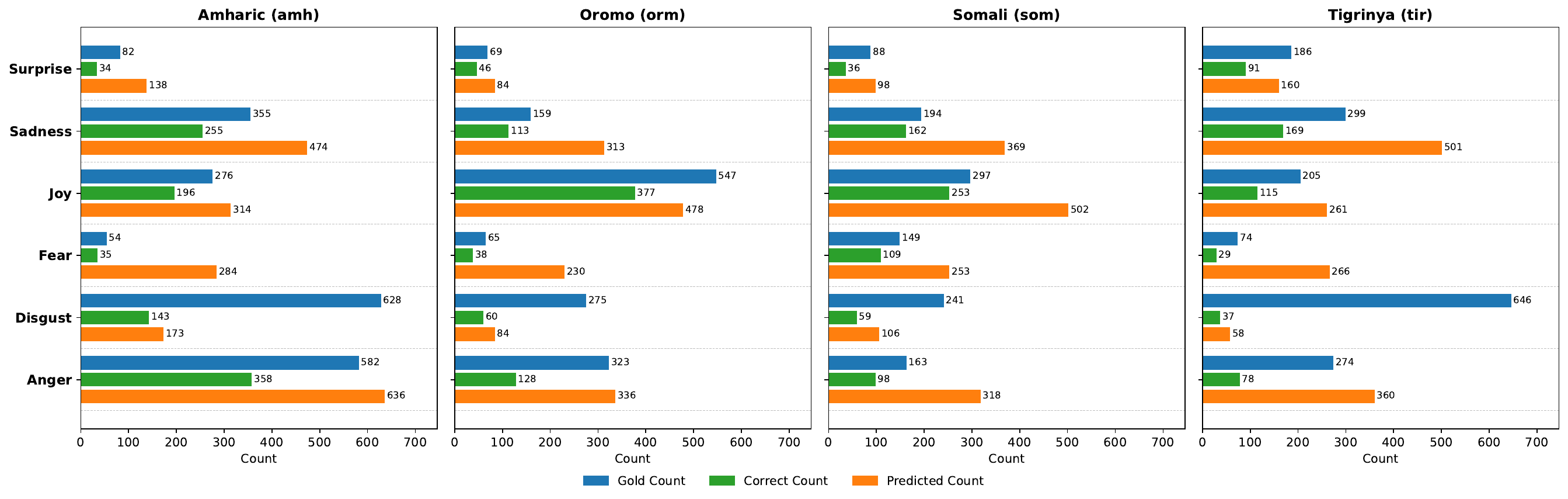}
    \caption{Emotion error analysis across languages and Emotion labels from the better LLMs (GPT-4.1-mini 5-shot). \textbf{Gold Count} is the number of human annotated labels, \textbf{Correct Count} is the labels the model predicted correctly from the total given gold count, and \textbf{Predicted Count} is the total predictions of the model for that specific emotion.}
    \label{fig:emotion-stat}
\end{figure*}

\subsection{Cross-lingual Emotion Classification}
\textbf{How does cross-lingual emotion transferability work among Ethiopian languages? }
Table \ref{tab:cross-lingual} reports the results of the cross-lingual transfer learning experiments. AfroXLMR-Social again achieves the highest performance for cross-lingual evaluations because it includes all the targeted Ethiopian languages during pretraining. Overall, when comparing cross-lingual results across \textbf{all multilingual} and \textbf{only similar-script} settings, all languages benefited from multilingual training. Specifically, Amharic (\texttt{amh}) and Tigrinya (\texttt{tir}) achieved higher transfer performance, likely because both use the Ethiopic script (Ge’ez). Among the BERT-like models, mBERT performs the worst, as none of the target languages was included during its pretraining. The AfroLM model performs the second worst, as it includes only Amharic (\texttt{amh}) in its pretraining.

\paragraph{Error Analysis}
We conducted a detailed prediction analysis for both emotion and intensity prediction to understand why LLMs (specifically GPT-4.1-mini 5-shot in this analysis) perform worse than BERT-like models. Figure \ref{fig:emotion-stat} presents detailed statistics, including the number of human-annotated labels (Gold Count), the number of correctly predicted labels (Correct Count) out of the total gold labels and the total number of predictions made by the model for each emotion (Predicted Count). Based on the statistics: 1) The number of correctly predicted labels is consistently lower than the number of gold labels, and 2) Mostly the model tends to over-predict, except for the \texttt{Disgust} emotion class across language.
Although the disgust emotion has one of the highest distributions across languages in the EthioEmo dataset, it is predicted less frequently than other emotion classes. A similar trend is observed in the intensity prediction results, where most errors are over-predictions, for example, assigning emotion intensities to instances that are not labeled with any emotion, and predicting 2 (medium) and 3 (high) intensities for instances that have low intensity annotations.

\section{Conclusion}
In this work, we extended the \textbf{EthioEmo} emotion dataset by adding the intensity of the corresponding labeled emotions. Using the dataset, we experiment with multi-label emotion classification, emotion intensity prediction, and cross-lingual transfer learning among Ethiopian languages. Generally, the African-centric language model such as AfroXLMR-Social that includes the evolution languages during pretraining performs is best for emotion, intensity, and cross-lingual emotion transferability between Ethiopian languages. This dataset and benchmark will contribute to the development of a more robust emotion evaluation task for low-resource languages. In future work,  we plan to release the annotator level data and suggest modeling the annotator level data instead of making the majority vote, as making the majority vote does not consider the minority perspectives of annotators for subjective NLP tasks such emotion analysis and emotion intensity prediction.

\section*{Limitations} Our work is not without limitation and we identified the following limitations with the future directions.
\paragraph{Limited Annotators per instance} While it is common to annotate multi-label emotion using three raters per instance, such as the GoEmotions dataset \cite{demszky-etal-2020-goemotions}, WRIME emotion intensity \cite{kajiwara-etal-2021-wrime}, and others, it is recommended that the more annotators, the higher the quality of the dataset \cite{troiano-etal-2021-emotion,suzuki-etal-2022-emotional}. For instance, the BRIGHTER emotion \cite{bright-muhammad2025} dataset intensity of the corresponding emotion is annotated by a minimum of five annotators. Based on our scope, we annotate the intensity using only a minimum of three raters per instance, and \texttt{amh} has five annotators per instance. Future work can add more annotations on top of our three annotators for the more quality of emotion intensity.
 
\paragraph{Majority vote limitation} Regarding deciding the final intensity, we determined the intensity label using majority vote and threshold average of the intensity values. This approach may not incorporate all the perspectives of annotators, as it is the general drawback of the majority vote. We plan to make the annotator-level data publicly available and open it for further exploration to determine the final emotion intensity. Alternatively, it can be used to model annotator perspectives without applying a majority vote. 

\paragraph{Limited models evaluation} We evaluated limited LLMs in a zero-shot setup based on our resource limitations. This evaluation can be extended by including more open-source LLMs, closed-source LLMs, and various few-shot evaluation setup. 
Additionally, the data can be explored in many ways such as if certain emotions or languages are more subjective than others.

\section*{Ethical Considerations} 
As started from a previously published dataset \cite{belay-etal-2025-evaluating}, emotion intensity annotation, perception, and expression are subjective and nuanced as they are strongly related to sociodemographic aspects (e.g., cultural background, social group, personal experiences, social context). Thus, we can never truly identify how one is feeling based solely on the given text snippets with absolute certainty. We ensure fair and honest analysis while conducting our work ethically and without harming anybody. 
\section{Bibliographical References}\label{sec:reference}
\bibliographystyle{lrec2026-natbib}
\bibliography{custom}


\onecolumn
\appendix
\section*{Appendix}\label{sec:appendix}

\section{Model Training/Testing Parameters}\label{app:parms}
Fine-tuning hyperparameters of encoder-only PLMs are epoch 3, lrate = 5e-5, max-token 256, and batch size 8.

\paragraph{Multi-label emotion classification prompt:} \texttt{"Evaluate whether the author of the following text conveys the emotion \{\{EMOTION\}\}. Think step by step before you answer. Answer with ONLY 'yes' or 'no'.
Do not provide any explanation.
\\ Text: \{text\} "}

\paragraph{Emotion intensity prediction prompt:}
\texttt{Determine the intensity of the emotions \{\{EMOTION\}\} in the text.
The intensity score ranges from 0 to 3: \\
- 0 = No intensity (emotion not present) \\
- 1 = Low intensity \\
- 2 = Medium intensity \\
- 3 = High intensity \\
Example output: \\
\{\{"anger": 0, "disgust": 2, "fear": 0, "joy": 1, "sadness": 0, "surprise": 0\}\} \\
Text: \{text\}}

\section{Additional Results}\label{app:more}
Table \ref{tab:class-result} presents detail results at each emotion class level from the best AfroXLMR-Social \cite{belay2025afroxlmr} model.
\begin{table*}[!h]
\centering
\resizebox{\textwidth}{!}{%
\begin{tabular}{lcccc|cccc}
\toprule
&\multicolumn{4}{c|}{\textbf{Emotion results (F1)}} & \multicolumn{4}{c}{\textbf{Intensity results (F1)}}\\
\cmidrule{2-5}\cmidrule{6-9}
\textbf{Models}& amh& orm & som & tir & amh& orm & som & tir\\
\midrule
Anger       &70.29&56.97&48.29&41.32     &47.35&27.98&13.81&00.00\\
Disgust     &79.37&64.21&55.31&77.46     &69.72&48.60&38.50&67.18\\
Fear        &57.78&38.25&56.30&43.52     &00.00&00.00&53.60&00.00\\
Joy         &75.91&82.64&66.85&64.66     &78.58&67.71&57.81&59.32\\
Sadness     &73.30&56.12&68.43&65.79     &80.16&49.30&66.91&60.23\\
Surprise    &67.34&66.22&33.33&68.67     &47.63&00.00&00.00&66.32\\
\midrule
Average     &70.66&60.74&54.75&60.24     &53.82&32.26&38.44&42.18\\
\bottomrule
\end{tabular}
}
\caption{Emotion class-level results from the best model (AfroXLML-Social).}
\label{tab:class-result}
\end{table*}

\section{Emotion Co-occurrence}
\begin{figure*}[!ht]
    \centering
    \includegraphics[width=\linewidth]{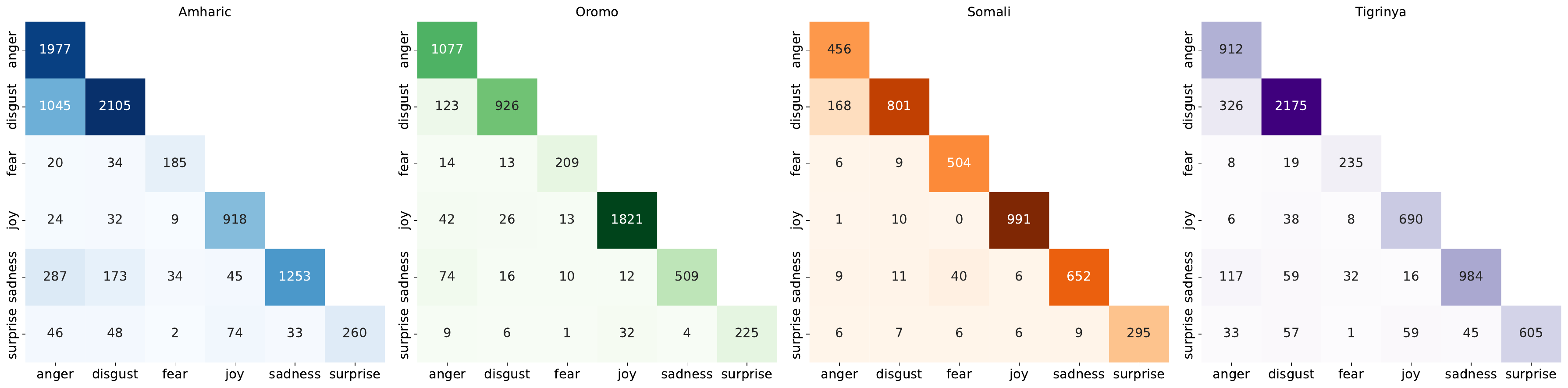}
    \caption{ Emotion co-occurrence across the six basic emotions and languages }
    \label{fig:co-occur}
\end{figure*}

Figure \ref{fig:co-occur} shows the co-occurrence between emotion classes.  Consistently in all languages, anger and disgust are the most common emotions that appear together. Anger, disgust, and joy are the top three emotions with the highest intensity level, as they also have the most statistics among other emotions, such as fear and surprise.

\section{Emotion label distribution}\label{app:label-dist}
\begin{figure*}[!ht]
     \centering
    \includegraphics[width=0.7\textwidth]{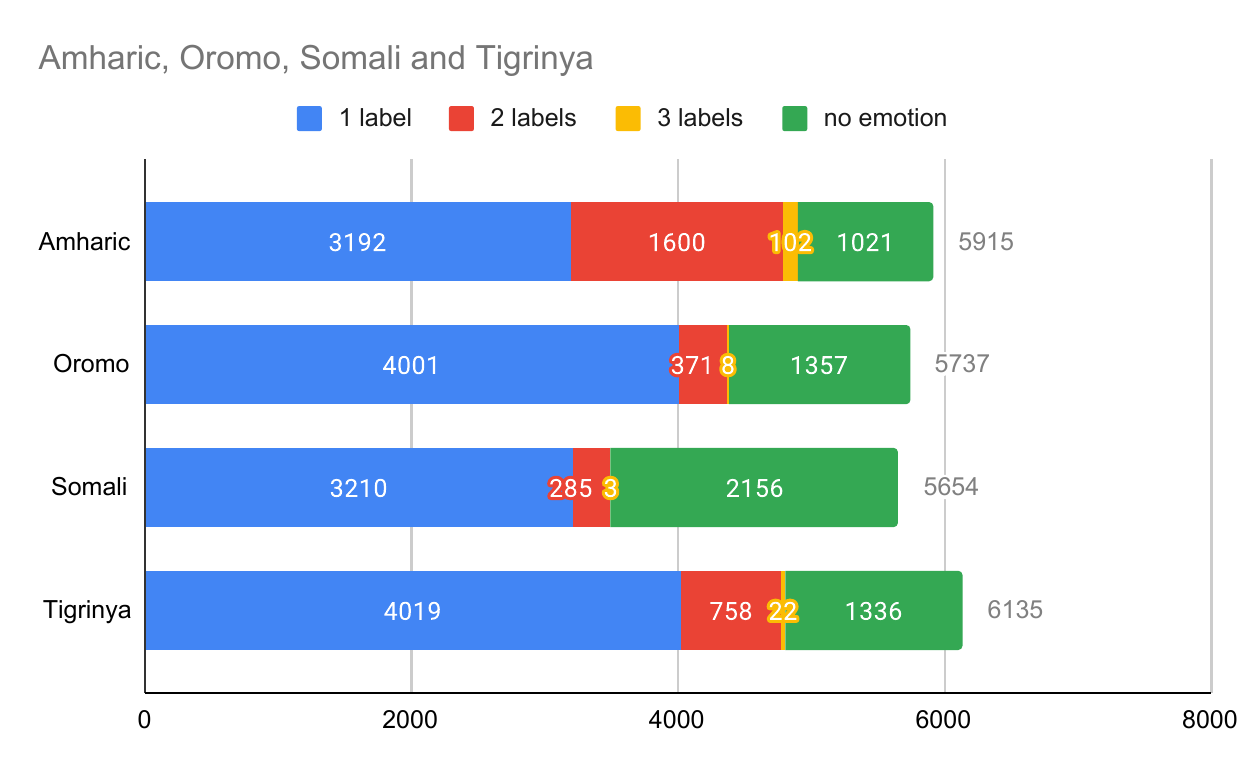}
    \caption{Emotion statistics in the number of emotion labels for each instance. Most of the instances in the dataset have a single emotion label. Amharic (1600) and Tigrinya (758) have a comparatively higher instance of double emotions than Oromo (371) and Somali (285).}
    \label{fig:stat}
\end{figure*}

Figure \ref{fig:stat} shows the emotion label distribution across languages. As EthioEmo is annotated in a multi-label emotion approach, a text might have no emotion, one, two, multiple, or all emotion labels; most instances across all languages have a single emotion label.  

\clearpage
\section{Emotion intensities prediction analysis}\label{app:intnsity}
Table \ref{app:intes-dist} shows the prediction distributions from GPT (GPT-4.1-mini 5-shot) intensity scales from 0 (none) to 3 (high) for the corresponding emotion and across languages. 
\begin{table*}[ht!]
\centering
\scriptsize
\setlength{\tabcolsep}{3pt}
\renewcommand{\arraystretch}{1.2}
\begin{tabular}{llrrrrrrrrrrrrrrrr}
\hline
\multirow{2}{*}{\textbf{Emotion}} & \multirow{2}{*}{\textbf{Intensity}} &
\multicolumn{3}{c}{\textbf{Amharic}} & &
\multicolumn{3}{c}{\textbf{Oromo}} & &
\multicolumn{3}{c}{\textbf{Somali}} & &
\multicolumn{3}{c}{\textbf{Tigrinya}} \\
\cline{3-5} \cline{7-9} \cline{11-13} \cline{15-17}
 & & Gold & Correct & Pred && Gold & Correct & Pred && Gold & Correct & Pred && Gold & Correct & Pred \\
\hline
Anger    & 0 & 1674 & 819 & 982 && 1436 & 788 & 853 && 1364 & 703 & 774 && 1566 & 493 & 535 \\
         & 1 & 54 & 23 & 723 && 41 & 18 & 447 && 41 & 16 & 456 && 49 & 18 & 651 \\
         & 2 & 166 & 47 & 326 && 98 & 30 & 322 && 147 & 37 & 355 && 179 & 39 & 332 \\
         & 3 & 41 & 12 & 307 && 31 & 13 & 283 && 25 & 7 & 264 && 46 & 15 & 322 \\
\hline
Disgust  & 0 & 1288 & 1043 & 1348 && 1139 & 1004 & 1308 && 1212 & 981 & 1226 && 1194 & 946 & 1304 \\
         & 1 & 69 & 22 & 410 && 72 & 23 & 414 && 65 & 24 & 394 && 68 & 21 & 409 \\
         & 2 & 391 & 45 & 120 && 387 & 56 & 129 && 395 & 59 & 112 && 398 & 47 & 118 \\
         & 3 & 171 & 6 & 9 && 162 & 9 & 8 && 176 & 8 & 8 && 180 & 7 & 9 \\
\hline
Fear     & 0 & 1690 & 1201 & 1192 && 1750 & 1153 & 1225 && 1774 & 1174 & 1186 && 1766 & 1162 & 1195 \\
         & 1 & 12 & 3 & 366 && 14 & 3 & 347 && 12 & 3 & 355 && 10 & 2 & 366 \\
         & 2 & 67 & 15 & 229 && 56 & 13 & 237 && 61 & 17 & 241 && 57 & 14 & 231 \\
         & 3 & 8 & 3 & 49 && 6 & 2 & 44 && 8 & 3 & 45 && 7 & 4 & 48 \\
\hline
Joy      & 0 & 1588 & 1334 & 1409 && 1689 & 1351 & 1447 && 1627 & 1367 & 1389 && 1635 & 1345 & 1413 \\
         & 1 & 25 & 6 & 179 && 27 & 7 & 183 && 24 & 6 & 172 && 26 & 5 & 173 \\
         & 2 & 136 & 30 & 167 && 129 & 32 & 175 && 138 & 33 & 168 && 134 & 28 & 169 \\
         & 3 & 41 & 16 & 86 && 43 & 15 & 82 && 44 & 17 & 88 && 45 & 15 & 85 \\
\hline
Sadness  & 0 & 1495 & 704 & 724 && 1534 & 694 & 722 && 1527 & 671 & 732 && 1541 & 655 & 721 \\
         & 1 & 16 & 5 & 530 && 17 & 4 & 527 && 20 & 6 & 544 && 18 & 5 & 540 \\
         & 2 & 185 & 49 & 389 && 191 & 50 & 394 && 189 & 48 & 388 && 192 & 50 & 396 \\
         & 3 & 93 & 32 & 184 && 84 & 29 & 183 && 87 & 27 & 186 && 89 & 31 & 183 \\
\hline
Surprise & 0 & 1662 & 1437 & 1503 && 1655 & 1433 & 1497 && 1648 & 1441 & 1516 && 1654 & 1429 & 1502 \\
         & 1 & 12 & 4 & 284 && 13 & 5 & 281 && 10 & 4 & 277 && 11 & 4 & 283 \\
         & 2 & 135 & 32 & 57 && 138 & 30 & 58 && 136 & 33 & 55 && 137 & 31 & 55 \\
         & 3 & 39 & 0 & 0 && 36 & 0 & 0 && 37 & 0 & 0 && 38 & 0 & 0 \\
\hline
\end{tabular}
\caption{Emotion intensity distribution across languages. The intensity levels are defined as 0 = no emotion, 1 = low, 2 = medium, and 3 = high for each corresponding emotion category. \textbf{Gold} denotes the number of human-annotated instances, \textbf{Correct} indicates the number of instances the model correctly predicted among the Gold labels, and \textbf{Pred.} (Prediction) represents the total number of instances the model predicted for that specific intensity level. Results are obtained from the best-performing LLM, GPT-4.1-mini, evaluated in a 5-shot setting.}
\label{tab:intensity-all}
\end{table*}

\clearpage
\section{Model details, papers, and its Hugging-face name}\label{app:model}
\begin{itemize}
    \item LaBSE \cite{feng-etal-2022-language} - sentence-transformers/LaBSE 
    \item RemBERT \cite{chung2020rembert} - google/rembert
    \item XLM-RoBERTa - FacebookAI/xlm-roberta-base (large) \cite{xlm-r}
    \item mDeBERTa \cite{he2021debertav3} - microsoft/mdeberta-v3-base
    \item mBERT \cite{libovický2019lmbert} - google-bert\_bert-base-multilingual-cased
    \item EthioLLM \cite{tonja-etal-2024-ethiollm} - EthioNLP/EthioLLM-l-70K : multilingual models for five Ethiopian languages (\texttt{amh}, gez, orm, \texttt{som}, and \texttt{tir}) and English.
    \item AfriBERTa \cite{ogueji-etal-2021-small} - castorini/afriberta\_large : pre-trained on 11 African languages. It includes our four target Ethiopian languages.
    \item AfroXLMR \cite{alabi-etal-2022-adapting} - Davlan/afro-xlmr-large-61L (76L) - adapted from XLM-R-large \cite{xlm-r} (has two versions: 61 and 76 languages) for African languages, including the four Ethiopian languages and high-resource languages such as English, French, Chinese, and Arabic.
    \item AfroLM \cite{dossou-etal-2022-afrolm} - bonadossou/afrolm\_active\_learning - a multilingual model pre-trained on 23 African languages, including \texttt{amh} and \texttt{orm} from Ethiopian languages. 
    \item AfroXLMR-Social \cite{belay2025afroxlmr} - Tadesse/AfroXLMR-Social - a multilingual model pre-trained on 23 African languages, including \texttt{amh} and \texttt{orm} from Ethiopian languages. 
    \item DeepSeek-R1-70 \cite{deepseekai2025} - deepseek-ai/DeepSeek-R1-Distill-Llama-70B
     \item Gemma-3-12B \cite{team2025gemma} - google/gemma-3-12b-it
    \item Llama-3.1-8B \cite{dubey2024llama} - meta-llama/Llama-3.1-8B-Instruct
    \item Llama-3.3-70B \cite{grattafiori2024llama} - meta-llama/Llama-3.3-70B-Instruct
\end{itemize}

\section{Annotation Guideline}\label{app:guideline}
Based on previous emotion and its corresponding intensity annotation guidelines for other languages, we prepared intensity annotation guidelines. We asked annotators as which of the options below best describes the feelings of the narrator (select one option for each row):
For the intensity annotation, we employed native speakers of each language. We provided detailed annotation guidelines with text examples, emotion label(s), and each intensity level of the emotions. We compensated annotators with an hourly wage in Ethiopia. A total of 20 males and five females participated in the annotation. Their academic status is a bachelor's degree or above. 

\begin{tabular}{l l l l}
0: No anger & 1: slight anger & 2: moderate anger & 3: high anger \\
0: No disgust & 1: slight disgust & 2: moderate disgust & 3: high disgust \\
0: No sadness & 1: slight sadness & 2: moderate sadness & 3: high sadness \\
0: No fear & 1: slight fear & 2: moderate fear & 3: high fear \\
0: No joy & 1: slight joy & 2: moderate joy & 3: high joy \\
0: No surprise & 1: slight surprise & 2: moderate surprise & 3: high surprise \\
\end{tabular}

\subsection{Emotion Categories and Definitions}
\textbf{Joy}: Expressions of happiness, pleasure, or contentment. Consider happiness to be a broad category that includes: joyful, elated, content, cheerful, blissful, delighted, gleeful, satisfied, ecstatic, upbeat, pleased, etc. "I just passed my exams! "\\
~\\ \textbf{Sadness}: Expressions of unhappiness, sorrow, or disappointment. Consider sadness to be a broad category that includes: melancholic, despondent, gloomy, heartbroken, longing, mourning, dejected, downcast, disheartened, dismayed, etc.
"I miss my family so much. It's been a tough year."\\
~\\ \textbf{Anger}: Expressions of frustration, irritation, or rage. Consider anger to be a broad category that includes: irritated, annoyed, aggravated, indignant, resentful, offended, exasperated, livid, irate. etc.
 "Why is the internet so slow today?!"\\
~\\ \textbf{Fear}: Expressions of anxiety, apprehension, or dread. Consider fear to be a broad category that includes: frightened, alarmed, apprehensive, intimidated, panicky, wary, dreadful, shaken, etc.
"There's a huge storm coming our way. I hope everyone stays safe."\\
~\\ \textbf{Surprise}: Expressions of astonishment or unexpected events. Consider surprise to be a broad category that includes: taken aback, bewildered,  astonished, amazed, startled, stunned, taken aback, shocked, dumbstruck, confounded, stupefied, etc.
 "I can't believe he just proposed to me!"\\
~\\ \textbf{Disgust}: A reaction to something offensive or unpleasant. Consider disgust to include distinguishing an individual/organization based solely on their identity/humanity, i.e., religion, ethnicity, language, and insulting, belittling, or obscene words. It means hating a person for his humanity.
 "I hate black people"

\paragraph{Summary Instructions}
Carefully read the detailed instructions before proceeding with the task. 
You will be given an {language} sentence taken randomly from a social media (X, video comments, and news headlines) and multiple choice options for the emotions the narrator is feeling. Mark all options that apply. 
The options correspond to seven emotions (anger, sadness, fear, disgust, happiness, and surprise) and select the intensity of the chosen emotion on a scale from no emotion, slight emotion, moderate emotion, to high emotion.

\paragraph{Quality Control}
Some questions have pre-determined correct answers. If you mark these questions incorrectly, we will give you immediate feedback in a pop-up box. An occasional misanswer is okay. However, if the rate of misanswering is high (e.g., >20\%), then all of one's HITs may be rejected.
\newpage
\subsection{Additional Results from LLMs}
Table \ref{tab:llm-additional} shows additional LLMs results from 0-shot and 5-shot. Across all tested languages, the Gemma-3 models consistently outperform the competition, with the 12b variant achieving the highest scores in both 0-shot and 5-shot settings. While most models show performance gains when moving from 0-shot to 5-shot prompts, others, such as LLaMa-3.1 and Ministral-8b, occasionally exhibit performance degradation or stagnation, suggesting a struggle with in-context learning in these low-resource languages.
\begin{table}[H]
\begin{tabular}{l|cc|cc|cc|cc|}
\toprule
 & \multicolumn{2}{c|}{amh} & \multicolumn{2}{c|}{orm} & \multicolumn{2}{c|}{som} & \multicolumn{2}{c|}{tir} \\
\textbf{Model} & 0 shot & 5 shot & 0 shot & 5 shot & 0 shot & 5 shot & 0 shot & 5 shot \\
\midrule
\href{https://huggingface.co/google/gemma-3-12b-it}{Gemma-3-12b-it} & 52.94 & 53.08 & 32.48 & 36.77 & 41.00 & 46.60 & 41.78 & 39.39 \\
\href{https://huggingface.co/google/gemma-3-4b-it}{Gemma-3-4b-it} & 45.31 & 47.88 & 19.78 & 22.35 & 33.60 & 38.35 & 30.81 & 36.67 \\
\href{https://huggingface.co/meta-llama/Llama-3.1-8B-Instruct}{LLaMa-3.1-8b-instruct} & 26.00 & 20.41 & 19.55 & 20.86 & 20.28 & 21.56 & 14.52 & 14.12 \\
\href{https://huggingface.co/meta-llama/Llama-3.2-3B-Instruct}{LLaMa-3.2-3b-instruct} & 6.61 & 15.56 & 5.85 & 12.19 & 9.94 & 13.59 & 4.71 & 12.27 \\
\href{https://huggingface.co/ministral/Ministral-3b-instruct}{Ministral-3b-instruct} & 0.00 & 0.37 & 0.00 & 6.19 & 0.00 & 6.08 & 0.00 & 0.00 \\
\href{https://huggingface.co/mistralai/Ministral-8B-Instruct-2410}{Ministral-8b-instruct-2410} & 16.90 & 10.81 & 16.16 & 14.96 & 14.26 & 14.84 & 12.40 & 9.85 \\
\href{https://huggingface.co/Qwen/Qwen2.5-3B-Instruct}{Qwen2.5-3B-Instruct} & 10.93 & 11.00 & 10.21 & 10.32 & 11.51 & 11.22 & 10.17 & 12.73 \\
\href{https://huggingface.co/Qwen/Qwen2.5-7B-Instruct}{Qwen2.5-7B-Instruct} & 16.32 & 21.24 & 1.68 & 15.03 & 2.45 & 20.59 & 14.40 & 16.80 \\
\bottomrule
\end{tabular}
\caption{Additional results from the LLMs' 0-shot and 5-shot.}
\label{tab:llm-additional}
\end{table}

\end{document}